%% file: main.tex
\documentclass[11pt,a4paper]{article}
\usepackage[hyperref]{emnlp-ijcnlp-2019}
\usepackage{times}
\usepackage{latexsym}
\usepackage{mathtools}
\usepackage{booktabs}
\usepackage{bm}
\usepackage{bbm}
\usepackage{soul}
\usepackage{colortbl}
\usepackage{amsmath}
\usepackage{amsfonts}
\usepackage{fancybox}
\usepackage{arydshln}
\usepackage{float}
\usepackage{multirow}
\usepackage{subfig}
\usepackage{url}
\usepackage{enumitem}

\aclfinalcopy 

\title{Sentence-Level Content Planning and Style Specification\\for Neural Text Generation}

\author{Xinyu Hua \and Lu Wang \\
  Khoury College of Computer Sciences \\
  Northeastern University \\
  Boston, MA 02115 \\
  {\tt hua.x@husky.neu.edu} \quad {\tt luwang@ccs.neu.edu}}

\date{}

\begin{document}
\maketitle
\begin{abstract}
\input{abstract.tex}
\end{abstract}

\section{Introduction}
\label{sec:intro}
\input{intro.tex}

\section{Related Work}
\input{related.tex}

\section{Model}
\input{model.tex}

\section{Tasks and Datasets}
\label{sec:dataset}
\input{data.tex}

\section{Experiments}
\input{experiment.tex}

\section{Results and Analysis}
\input{analysis.tex}

\section{Conclusion}
\input{conclusion.tex}

\section*{Acknowledgements}
This research is supported in part by National Science Foundation through Grants IIS-1566382 and
IIS-1813341, and Nvidia GPU gifts. We are grateful to Rik Koncel-Kedziorski and Hannaneh Hajishirzi for sharing their system outputs. 
We also thank anonymous reviewers for their valuable suggestions. 

\bibliography{ref}
\bibliographystyle{acl_natbib}

\appendix 
\section{Appendices}
\input{appendix.tex}
\input{tables.tex}

\end{document}

%% file: abstract.tex
Building effective text generation systems requires three critical components: content selection, text planning, and surface realization, and traditionally they are tackled as separate problems. 
Recent all-in-one style neural generation models have made impressive progress, yet they often produce outputs that are incoherent and unfaithful to the input. 
To address these issues, we present an end-to-end trained two-step generation model, where a sentence-level content planner first decides on the keyphrases to cover as well as a desired language style, followed by a surface realization decoder that generates relevant and coherent text. 
For experiments, we consider three tasks from domains with diverse topics and varying language styles: persuasive argument construction from Reddit, paragraph generation for normal and simple versions of Wikipedia, and abstract generation for scientific articles.
Automatic evaluation shows that our system can significantly outperform competitive comparisons. 
Human judges further rate our system generated text as more fluent and correct, compared to the generations by its variants that do not consider language style.

%% file: intro.tex
Automatic text generation is a long-standing challenging task, as it needs to solve at least three major problems: (1) {\it content selection} (``what to say"), identifying pertinent information to present, (2) {\it text planning} (``when to say what"), arranging content into ordered sentences, and (3) {\it surface realization} (``how to say it"), deciding words and syntactic structures that deliver a coherent output based on given discourse goals~\cite{mckeown1992text}. 
Traditional text generation systems often handle each component separately, thus requiring extensive effort on data acquisition and system engineering~\cite{reiter2000building}. 
Recent progress has been made by developing end-to-end trained neural models~\cite{rush-etal-2015-neural,yu-etal-2018-neural,fan-etal-2018-hierarchical}, which naturally excel at producing fluent text.
Nonetheless, limitations of model structures and training objectives make them suffer from low interpretability and substandard generations which are often incoherent and unfaithful to the input material~\cite{see-etal-2017-get,wiseman-etal-2017-challenges,li-etal-2017-adversarial}. 

\begin{figure}[t]
    \hspace{-2mm}
    \centering
    \includegraphics[width=78mm]{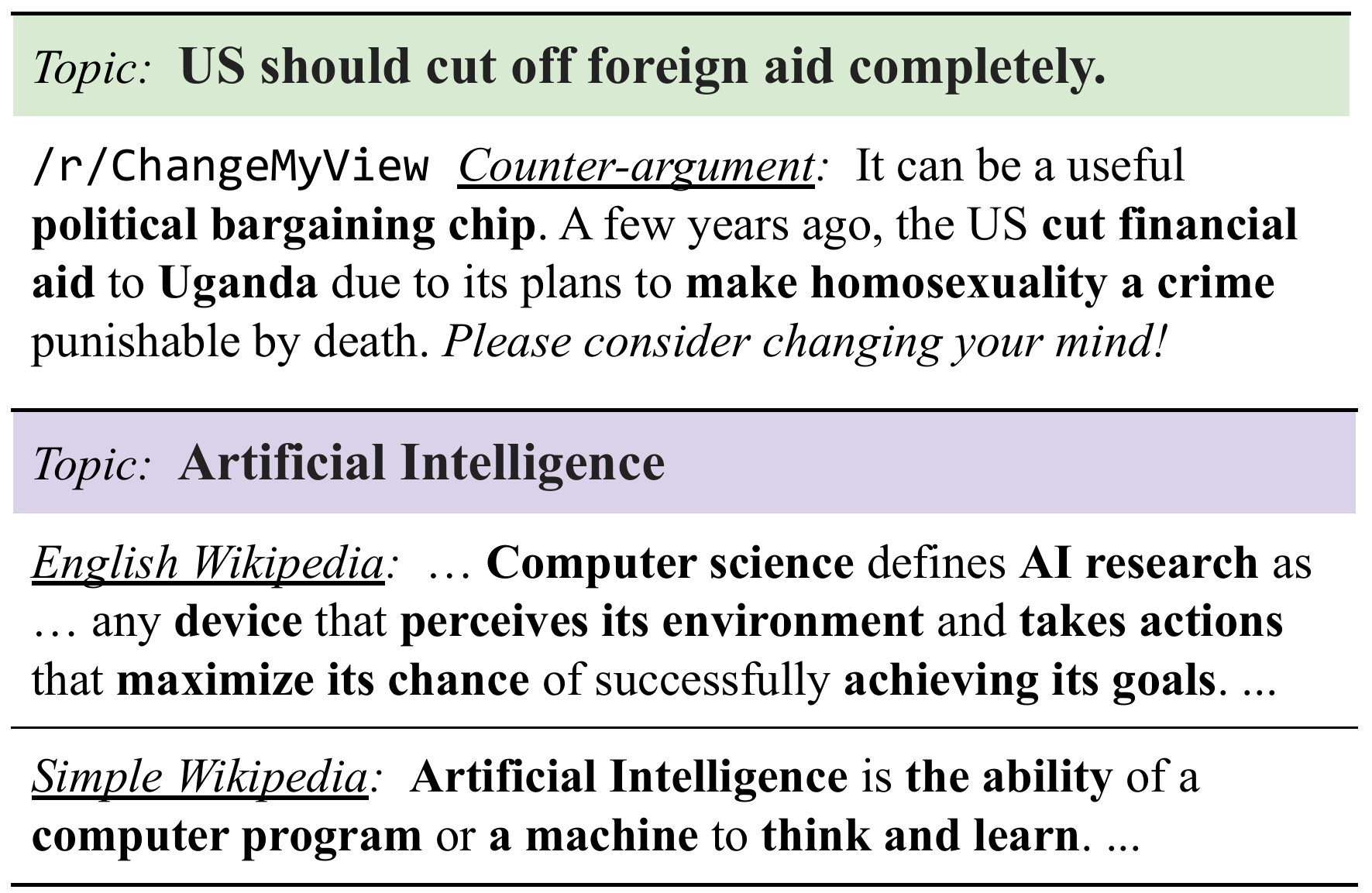}
    \caption{
    [Upper] Sample counter-argument from Reddit. Argumentative stylistic language for persuasion is in italics. 
    [Bottom] Excerpts from Wikipedia, where sophisticated concepts and language of higher complexity used in the standard version are not present in the corresponding simplified version. 
    Both: key concepts are in bold. 
    }
    \label{fig:motivating_example}
\end{figure}

\begin{figure*}[t]
\centering
  \includegraphics[height=52mm]{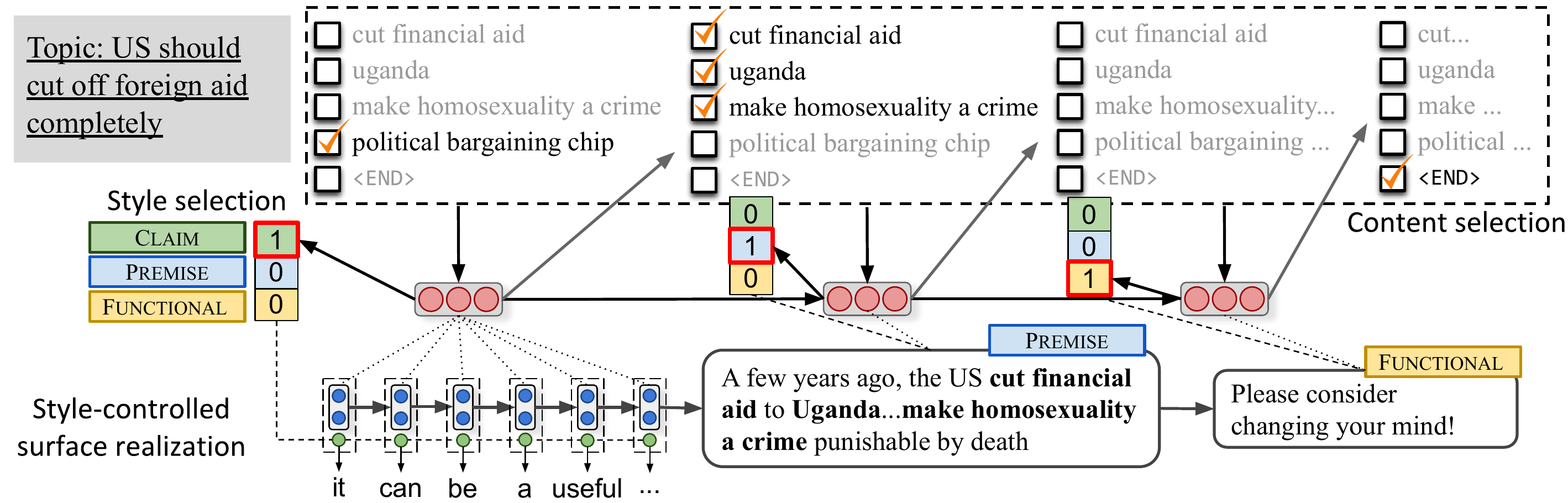}
  \caption{
    Overview of our framework. The LSTM content planning decoder (\S~\ref{subsec:planning}) first identifies a set of keyphrases from the memory bank conditional on previous selection history, based on which, a style is specified. 
    During surface realization, the hidden states of the planning decoder and the predicted style encoding are fed into the realizer, which generates the final output (\S~\ref{subsec:realization}). Best viewed in color.  
  }
  \label{fig:pipeline}
\end{figure*}

To address the problems, we believe it is imperative for neural models to gain adequate control on content planning (i.e., content selection and ordering) to produce coherent output, especially for long text generation. 
We further argue that, in order to achieve desired discourse goals, it is beneficial to enable style-controlled surface realization by explicitly modeling and specifying proper linguistic styles.
Consider the task of producing counter-arguments to the topic ``US should cut off foreign aid completely". A sample argument in Figure~\ref{fig:motivating_example} demonstrates how human selects a series of talking points and a proper style based on the argumentative function for each sentence. For instance, the argument starts with a {\it proposition} on ``foreign aid as a political bargaining chip", followed by a concrete {\it example} covering several key concepts. It ends with {\it argumentative stylistic language}, which differs in both content and style from the previous sentences. 
Figure~\ref{fig:motivating_example} shows another example on Wikipedia articles: compared to a topic's standard version where longer sentences with complicated concepts are constructed, its simplified counterpart tends to explain the same subject with plain language and simpler concepts, indicating the interplay between content selection and language style.

We thus present an end-to-end trained neural text generation framework that includes the modeling of traditional generation components, to promote the control of content and linguistic style of the produced text.\footnote{Data and code are available at \url{xinyuhua.github.io/Resources/emnlp19/}.}
Our model performs {\it sentence-level content planning} for information selection and ordering, and {\it style-controlled surface realization} to produce the final generation.
We focus on conditional text generation problems~\cite{lebret-etal-2016-neural,colin-etal-2016-webnlg,dusek-etal-2018-findings}: As shown in Figure~\ref{fig:pipeline}, the input to our model consists of a topic statement and a set of keyphrases. The output is a relevant and coherent paragraph to reflect the salient points from the input. 
We utilize two separate decoders: for each sentence, (1) a planning decoder selects relevant keyphrases and a desired style conditional on previous selections, and (2) a realization decoder produces the text in the specified style. 

{We demonstrate the effectiveness of our framework on three challenging datasets with diverse topics and varying linguistic styles: persuasive argument generation on Reddit ChangeMyView~\cite{hua-wang-2018-neural}; introduction paragraph generation on a {\it newly} collected dataset from Wikipedia and its simple version; and scientific paper abstract generation on AGENDA dataset~\cite{koncel-kedziorski-etal-2019-text}.}

Experimental results on all three datasets show that our models that consider content planning and style selection achieve significantly better BLEU, ROUGE, and METEOR scores than non-trivial comparisons that do not consider such information. Human judges also rate our model generations as more fluent and correct compared to the outputs produced by its variants without style modeling.

%% file: related.tex
Content selection and text planning are critical components in traditional text generation systems~\cite{reiter2000building}. 
Early approaches separately construct each module and mainly rely on hand-crafted rules based on discourse theory~\cite{scott1990getting, hovy1993automated} and expert knowledge~\cite{reiter-etal-2000-knowledge}, or train statistical classifiers with rich features~\cite{duboue-mckeown-2003-statistical,barzilay-lapata-2005-collective}. 
Advances in neural generation models have alleviated human efforts on system engineering, by combining all components into an end-to-end trained conditional text generation framework~\cite{mei-etal-2016-talk,wiseman-etal-2017-challenges}. However, without proper planning and control~\cite{rambow-korelsky-1992-applied,stone-doran-1997-sentence-planning,walker-etal-2001-spot}, the outputs are often found to be incoherent and hallucinating. 
Recent work~\cite{moryossef-etal-2019-step} separates content selection from the neural generation process and shows improved generation quality. However, their method requires an exhaustive search for content ordering and is therefore hard to generalize and scale. 
In this work, we improve the content selection by incorporating past selection history and directly feeding the predicted language style into the realization module.

Our work is also inline with concept-to-text generation, where sentences are produced from {\it structured} representations, such as database records~\cite{konstas-lapata-2013-inducing, lebret-etal-2016-neural, wiseman-etal-2017-challenges, moryossef-etal-2019-step}, knowledge base items~\cite{luan-etal-2018-multi,koncel-kedziorski-etal-2019-text}, and AMR graphs~\cite{konstas-etal-2017-neural, song-etal-2018-graph, koncel-kedziorski-etal-2019-text}. 
Shared tasks such as WebNLG~\cite{colin-etal-2016-webnlg} and E2E NLG challenges~\cite{dusek2019e2e} have been designed to evaluate single sentence planning and realization from the given structured inputs with a small set of fixed attribute types. 
Planning for multiple sentences in the same paragraph is nevertheless much less studied; it poses extra challenges for generating coherent long text, which is addressed in this work. 
Moreover, structured inputs are only available in a limited number of domains~\cite{tanaka1998reactive,chen2008learning,belz2008automatic,liang-etal-2009-learning,chisholm-etal-2017-learning}.  
The emerging trend is to explore less structured data~\cite{kiddon-etal-2016-globally,fan-etal-2018-hierarchical,martin2018event}. 
In our work, keyphrases are used as input to our generation system, which offer flexibility for concept representation and generalizability to broader domains. 

%% file: model.tex
Our model tackles conditional text generation tasks where the input is comprised of two major parts: (1) a topic statement, $\mathbf{x}=\{x_i\}$, which can be an argument, the title of a Wikipedia article, or a scientific paper title, and (2) a keyphrase memory bank, $\mathcal{M}$, containing a list of talking points, which plays a critical role in content planning and style selection. We aim to produce a sequence of words, $\mathbf{y}=\{y_t\}$, to comprise the output, which can be a counter-argument, a paragraph as in Wikipedia articles, or a paper abstract.

\subsection{Input Encoding}
\label{subsec:encoder}
The input text $\mathbf{x}$ is encoded via a bidirectional LSTM (biLSTM), with its last hidden state used as the initial states for both content planning decoder and surface realization decoder. 
To encode keyphrases in the memory bank $\mathcal{M}$, each keyphrase is first converted into a vector $\bm{e}_k$ by summing up all its words' embeddings from GloVe~\cite{pennington-etal-2014-glove}. A biLSTM-based {\it keyphrase reader}, with hidden states $\bm{h}^e_k$, is used to encode all keyphrases in $\mathcal{M}$. 
We also insert entries of \texttt{<START>} and \texttt{<END>} into $\mathcal{M}$ to facilitate learning to start and finish selection.

\subsection{Sentence-Level Content Planning and Style Specification}
\label{subsec:planning}

\noindent \textbf{Content Planning: Context-Aware Keyphrase Selection.}
Our content planner selects a set of keyphrases from the memory bank $\mathcal{M}$ for each sentence, indexed with $j$, conditional on keyphrases that have been selected in previous sentences, allowing topical coherence and content repetition avoidance. 
The decisions are denoted as a {\it selection vector} $\bm{v}_{j}\in\mathbb{R}^{|\mathcal{M}|}$, with each dimension $\bm{v}_{j,k}\in\{0,1\}$, indicating whether the $k$-th phrase is selected for the $j$-th sentence generation. 
Starting with a \texttt{<START>} tag as the input for the first step, our planner predicts $\bm{v}_{1}$ for the first sentence, and recurrently makes predictions per sentence until \texttt{<END>} is selected, as depicted in Figure~\ref{fig:pipeline}. 

Formally, we utilize a sentence-level LSTM $f$, which consumes the summation embedding of selected keyphrases, $\bm{m}_j$, to produce a hidden state $\bm{s}_j$ for the $j$-th sentence step:

{\fontsize{10}{11}\selectfont
\setlength{\abovedisplayskip}{2pt}
\setlength{\belowdisplayskip}{2pt}
\begin{align}
 	 & \bm{s}_{j} = f(\bm{s}_{j-1}, \bm{m}_j) \label{eq:sp-states} \\
 	 & \bm{m}_j = \sum_{k = 1}^{\mathcal{|M|}}{{
 	 {\bm{v}}}_{j, k}}\bm{h}^e_k \label{eq:sp-input}
\end{align}
}
where ${\bm{v}}_{j, k}\in\{0, 1\}$ is the selection decision for the $k$-th keyphrase in the $j$-th sentence.

Our recent work~\cite{hua-etal-2019-argument-generation} utilizes a similar formulation for sentence representations. However, the prediction of ${\bm{v}}_{j+1}$ is estimated by a bilinear product between $\bm{h}^e_k$ and $\bm{s}_j$, which is agnostic to what have been selected so far. While in reality, content selection for a new sentence should depend on previous selections. For instance, keyphrases that have already been utilized many times are less likely to be picked again; topically related concepts tend to be mentioned closely. We therefore propose a vector $\bm{q}_j$ that keeps track of what keyphrases have been selected up to the $j$-th sentence: 


{\fontsize{10}{11}\selectfont
\setlength{\abovedisplayskip}{2pt}
\setlength{\belowdisplayskip}{2pt}
\begin{align}
 	&\bm{q}_j = (\sum_{r=0}^j{\bm{v}_r})^T\times \mathbb{E} 
\end{align}
}
where $\mathbb{E} =  \lbrack \bm{h}^e_1, \bm{h}^e_2, \ldots \bm{h}^e_{|\mathcal{M}|} \rbrack^T  \in \mathbb{R}^{|\mathcal{M}|\times H}$ is the matrix of keyphrase representations, $H$ is the hidden dimension of the keyphrase reader LSTM.

Then $\bm{v}_{j+1}$ is calculated in an attentive manner with $\bm{q}_j$ as the attention query:

{\fontsize{10}{11}\selectfont
\setlength{\abovedisplayskip}{2pt}
\setlength{\belowdisplayskip}{2pt}
\begin{align}
 	&P({\bm{v}}_{j+1,k}=1|\bm{v}_{1:j}) = 
 	\sigma(\mathbf{w}^{T}_{v}\bm{s}_j + \bm{q}_j\mathbf{W}^{c}\bm{h}^e_{k}) \label{eq:kp-pred}
\end{align}
}
where $\sigma$ is the sigmoid funciton, and $\mathbf{w}_{\ast}$, $\mathbf{W}^{\ast}$, and $\mathbf{W}^{\ast \ast}$ are trainable parameters throughout the paper. Bias terms are all omitted for simplicity.

As part of the learning objective, we utilize the binary cross-entropy loss with the gold-standard selection $\bm{v}^*_j$ as criterion over the training set $D$: 

{\fontsize{10}{11}\selectfont
\setlength{\abovedisplayskip}{2pt}
\setlength{\belowdisplayskip}{2pt}
\begin{align}
\begin{split}
  & \mathcal{L}_{\text{sel}}= 
  - \sum_{(\mathbf{x}, \mathbf{y})\in D}\sum_{j=1}^{J}(\sum_{k=1}^{|\mathcal{M}|} \log(P(\bm{v}^*_{j,k})))\label{eq:l_sel} \\
\end{split}
\end{align}
}

\noindent \textbf{Style Specification.}
As discussed in \S~\ref{sec:intro}, depending on the content (represented as selected keyphrases in our model), humans often choose different language styles adapted for different discourse goals. 
Our model characterizes such stylistic variations by assigning a categorical style type $\bm{t}_j$ for each sentence, which is predicted as follows:

{\fontsize{10}{11}\selectfont
\setlength{\abovedisplayskip}{2pt}
\setlength{\belowdisplayskip}{2pt}
\begin{align}
 	 & \hat{{\bm{t}}}_j = \text{softmax}(\mathbf{w}_s^T(\tanh{(\mathbf{W}^{s}[\bm{m}_j;\bm{s}_j])}) \label{eq:type-pred} 
\end{align}
}

$\hat{\bm{t}}_j$ is the estimated distribution over all types. We select the one with the highest probability and use a one-hot encoding vector, $\bm{t}_j$, as the input to our realization decoder (\S~\ref{subsec:realization}). 
The estimated distributions $\hat{\bm{t}}_j$ are compared against the gold-standard labels $\bm{t}^*_j$ to calculate the cross-entropy loss $\mathcal{L}_{\text{style}}$:

\vspace{-4mm}
{\fontsize{10}{11}\selectfont
\setlength{\abovedisplayskip}{2pt}
\setlength{\belowdisplayskip}{2pt}
\begin{align}
 	& \mathcal{L}_{\text{style}} = -{\sum_{(\mathbf{x}, \mathbf{y})\in D}}\sum_{j=1}^J \bm{t}^*_j \log  \hat{\bm{t}}_{j} \label{eq:l_style}
\end{align}
\vspace{-2mm}
}

\subsection{Style-Controlled Surface Realization}
\label{subsec:realization}

Our surface realization decoder is implemented with an LSTM with state calculation function $g$ to get each hidden state $\bm{z}_t$ for the $t$-th generated token. 
To compute $\bm{z}_t$, we incorporate the content planning decoder hidden state $\bm{s}_{J(t)}$ for the sentence to be generated, with $J(t)$ as the sentence index, and previously generated token $\bm{y}_{t-1}$: 

{\fontsize{10}{11}\selectfont
\setlength{\abovedisplayskip}{2pt}
\setlength{\belowdisplayskip}{2pt}
\begin{align}
\begin{split}
 	\bm{z}_t 
 	= g(\bm{z}_{t-1}, \tanh(\mathbf{W}^{ws}\bm{s}_{J(t)} + \mathbf{W}^{ww}\bm{y}_{t-1})) \label{eq:real-dec}\\
\end{split}
\end{align}
}

For word prediction, we calculate two attentions, one over the input statement $\mathbf{x}$, which produces a context vector $\bm{c}^w_t$ (Eq. \ref{eq:enc-attn}), the other over the keyphrase memory bank $\mathcal{M}$, which generates $\bm{c}^e_t$ (Eq. \ref{eq:kp-attn-soft}). 
To better reflect the control over word choice by language styles, we directly append the predicted style $\bm{t}_{J(t)}$ to the context vectors and hidden state $\bm{z}_t$, to compute the distribution over the vocabulary\footnote{The inclusion of style variables is different from our prior style-aware generation model~\cite{hua-etal-2019-argument-generation}, where styles are predicted but not encoded for word production.}: 

{\fontsize{10}{11}\selectfont
\setlength{\abovedisplayskip}{2pt}
\setlength{\belowdisplayskip}{2pt}
\begin{align}
\begin{split}
    P(y_t|y_{1:{t-1}}) &= \text{softmax}(\tanh({\mathbf{W}^o[\bm{z}_t;\bm{c}^w_t;\bm{c}^e_t;\bm{t}_{J(t)}]}))\label{eq:output-prob}  \\
\end{split} \\
\begin{split}
    \bm{c}^w_t &= \sum_{i=1}^{L}\alpha^w_i\bm{h}_i , \,\,\,\,\,\,  \alpha^w_i = \text{softmax}(\bm{z}_t\mathbf{W}^{wa}\bm{h}_i)  \label{eq:enc-attn}\\
\end{split} \\
\begin{split}
    \bm{c}^e_t &= \sum_{k=1}^{|\mathcal{M}|}\alpha_k\bm{h}^e_k , \,\,\,\,\,\, \alpha^e_k = \text{softmax}(\bm{z}_t\mathbf{W}^{we}\bm{h}^e_k) \label{eq:kp-attn-soft}\\
\end{split} 
\end{align}
}

We further adopt a copying mechanism from~\newcite{see-etal-2017-get} to enable direct reuse of words from the input $\mathbf{x}$ and keyphrase bank $\mathcal{M}$ to allow out-of-vocabulary words to be included.

\subsection{Training Objective}
\label{sec:objective}
We jointly learn to conduct content planning and surface realization by aggregating the losses over (i) word generation: {\fontsize{9}{11}\selectfont $\mathcal{L}_{\text{gen}}=-{\sum_{D}}\sum_{t=1}^T \log P(y^*_t|\mathbf{x};\theta)$}, (ii) keyphrase selection: {\fontsize{9}{11}\selectfont $\mathcal{L}_{\text{sel}}$} (Eq. \ref{eq:l_sel}), and (iii) style prediction {\fontsize{9}{11}\selectfont $\mathcal{L}_{\text{style}}$} (Eq. \ref{eq:l_style}):

{\fontsize{10}{11}\selectfont
\setlength{\abovedisplayskip}{2pt}
\setlength{\belowdisplayskip}{2pt}
\begin{align}
\begin{split}
  & \mathcal{L}(\theta) = \mathcal{L}_{\text{gen}}(\theta) + \gamma \cdot \mathcal{L}_{\text{style}}(\theta) + \eta \cdot \mathcal{L}_{\text{sel}}(\theta) \\
\end{split}
\end{align}
}
where $\theta$ denotes the trainable parameters. $\gamma$ and $\eta$ are set to $1.0$ in our experiments for simplicity. 

%% file: data.tex
\subsection{Task I: Argument Generation}
Our first task is to generate a counter-argument for a given statement on a controversial issue. The input keyphrases are extracted from automatically retrieved and reranked passages with queries constructed from the input statement.

{
We reuse the dataset from our previous work~\cite{hua-etal-2019-argument-generation}, but annotate with newly designed style scheme. We first briefly summarize the procedures for data collection, keyphrase extraction and selection, and passage reranking; more details can be found in our prior work. Then we describe how to label argument sentences with style types that capture argumentative structures. 
}

The dataset is collected from Reddit \texttt{/r/ChangeMyView} subcommunity, where each thread consists of a multi-paragraph original post (OP), followed by user replies with the intention to change the opinion of the OP user. 
Each OP is considered as the input, and the root replies awarded with delta ($\Delta$), or with positive \texttt{karma} (\# upvotes $>$ \# downvotes) are target counter-arguments to be generated. A domain classifier is further adopted to select politics related threads. 
Since users often have separate arguments in different paragraphs, we treat each paragraph as one {\it target argument} by itself. Statistics are shown in Table~\ref{tab:data-stat}. 

\begin{table}[t]
\fontsize{9}{11}\selectfont
 \setlength{\tabcolsep}{0.5mm}
  \centering
    \begin{tabular}{lccc}
    \toprule
    &  {\bf Argument } & {\bf Wikipedia}& {\bf AGENDA} \\
    &  \#~Args (\#~Threads) & (Nor. / Sim.) &\\
    \midrule
    \# Train & 272,147 (11,434) & 125,136 & 38,720 \\
    \# Dev & 40,291 (1,784) & 21,004 & 1,000 \\
    \# Test & 46,757 (1,706) & 23,534 &  1,000 \\
    {\# Tokens} & 54.87 & 70.57 / 48.60 & 141.34 \\
    \# Sent. & 2.48 & 3.15 / 3.20 & 5.59 \\
    \# {KP (candidates)} & 55.80 & 23.56 & 12.23 \\
    \# {KP (selected)} & 11.61 & 16.01/11.11 & 12.23 \\
    \bottomrule
    \end{tabular}
    \caption{
    Statistics of the three datasets. {Average numbers are reported. 
    For argument dataset, number of unique threads is also shown. {On AGENDA, entities are extracted from abstract as keyphrases, hence all candidates are ``selected". }}
    }
    \label{tab:data-stat}
    \vspace{-4mm}
\end{table}

\smallskip
\noindent \textbf{Input Keyphrases and Label Construction.} 
{To obtain the input keyphrase candidates and their sentence-level selection labels, we first construct queries to retrieve passages from Wikipedia and news articles collected from \url{commoncrawl.org}.\footnote{The choice of news portals, statistics of the dataset, and preprocerssing steps are described in \newcite{hua-etal-2019-argument-generation}, \S 4.1.} For training, we construct a query per target argument sentence using its content words for retrieval, and keep top $5$ passages per query. 
For testing, the queries are constructed from the sentences in OP (input statement). }

{We then extract keyphrases from the retrieved passages based on topic signature words~\cite{C00-1072} calculated over the given OP. These words, together with their related terms from WordNet~\cite{H94-1111}, are used to determine whether a phrase in the passage is a {\it keyphrase}. Specifically, a keyphrase is (1) a noun phrase or verb phrase that is shorter than 10 tokens; (2) contains at least one content word; (3) has a topic signature or a Wikipedia title. For each keyphrase candidate, we match them with the sentences in the target counter-argument, and we consider it to be ``selected" for the sentence if there is any overlapping content word.}

{During test time, we further adopt a stance classifier from \newcite{E17-1024} to produce a stance score for each passage. We retain passages that have a negative stance towards OP, and a greater than 5 stance score. They are further ordered based on the number of overlapping keyphrases with the OP. Top 10 passages are used to construct the input keyphrase bank, and as optional input to our model.}

\begin{table}[t]
\fontsize{10}{11}\selectfont
 \setlength{\tabcolsep}{0.7mm}
  \centering
  \begin{tabular}{lccc}
    \toprule
    &   \textsc{Claim} & \textsc{Premise} & \textsc{Functional} \\
    \midrule
    \# Arguments & 29.1\% & 62.2\% & 8.7\% \\
    \# Tokens & 17.0 & 26.2 & 10.0 \\
    \bottomrule
    \end{tabular}\vspace{2mm}
    \begin{tabular}{c ccccc}
    \toprule
    Length $\in$ &  (0, 10] & (10, 20] & (20, 30] & (30, $\infty)$ \\
    \midrule
    Normal Wikipedia & 9.9\% & 40.5\% & 29.8\% & 19.8\%  \\
    Simple Wikipedia & 29.3\% & 51.7\% & 14.6\% & 4.4\% \\
    \bottomrule
    \end{tabular}
    \caption{
     Sentence style distribution for argument and Wikipedia datasets. 
     }
    \label{tab:wikigen-type}
    \vspace{-2mm}
\end{table}

\smallskip
\noindent \textbf{Sentence Style Label Construction.} 
For argument generation, we define three sentence styles based on their argumentative discourse functions~\cite{persing-ng-2016-end,lippi2016argumentation}: 
\textsc{Claim} is a proposition, usually containing one or two talking points, e.g., ``{\it I believe foreign aid is a useful bargaining chip}"; 
\textsc{Premise} contains supporting arguments with reasoning or examples; 
\textsc{Functional} is usually a generic statement, e.g., ``{\it I understand what you said}". 
For training, we employ a list of rules extended from the claim detection method by \newcite{levy-etal-2018-towards} to automatically construct a style label for each sentence. Statistics are displayed in Table~\ref{tab:wikigen-type}, and sample rules are shown below, with the complete list in the Supplementary:

{\fontsize{10}{12}\selectfont
\begin{itemize}
    
    \item \textsc{Claim}: must be shorter than 20 tokens and matches any of the following patterns: 
    (a) \texttt{i (don't)? (believe|agree|\ldots);} 
    (b) \texttt{(anyone|all|everyone|nobody\ldots) (should|could|need|must|might\ldots)};
    (c) \texttt{(in my opinion|my view|\ldots)}
    
    \item \textsc{Premise}: must be longer than 5 tokens, contains at least one noun or verb content word, and matches any of the following patterns:
    (a) \texttt{(for (example|instance)|e.g.)};
    (b) \texttt{(increase|reduce|improve|\ldots)}
    
    \item \textsc{Functional}: contains fewer than 5 alphabetical words and no noun or verb content word
    
\end{itemize}
}

Paragraphs that only contain \textsc{Functional} sentences are removed from our dataset.

\subsection{Task II: Paragraph Generation for Normal and Simple Wikipedia}
The second task is generating introduction paragraphs for Wikipedia articles. 
The input consists of a title, a user-specified global style (normal or simple), and a list of keyphrases collected from the gold-standard paragraphs of both normal and simple Wikipedia. During training and testing, the global style is encoded as one extra bit appended to $\bm{m}_j$ (Eq. \ref{eq:sp-input}).

We construct a \textit{new} dataset with topically-aligned paragraphs from normal and simple English Wikipedia.\footnote{We download the dumps of 2019/04/01 for both dataset.}
For alignment, we consider it a match if two articles share exactly the same title with at most two non-English words. We then extract the first paragraphs from both and filter out the pair if one of the paragraphs is shorter than $10$ words or is followed by a table.

\smallskip
\noindent \textbf{Input Keyphrases and Label Construction.} 
Similar to argument generation, we extract noun phrases and verb phrases and consider the ones with at least one content word as keyphrase candidates. After de-duplication, there are on average $5.4$ and $3.7$ keyphrases per sentence for the normal and simple Wikipedia paragraphs, respectively. 
For each sample, we merge the keyphrases from the aligned paragraphs as the input. The model is then trained to select the appropriate ones conditioned on the global style.

\smallskip
\noindent \textbf{Sentence Style Label Construction.} 
We distinguish sentence-level styles based on language complexity, which is approximated by sentence length. The distribution of sentence styles is displayed in Table \ref{tab:wikigen-type}.

\subsection{Task III: Paper Abstract Generation}
We further consider a task of generating abstracts for scientific papers~\cite{ammar-etal-2018-construction}, where the input contains a paper title and scientific entities mentioned in the abstract. 
We use the AGENDA data processed by~\newcite{koncel-kedziorski-etal-2019-text}, where entities and their relations in the abstracts are extracted by SciIE~\cite{luan-etal-2018-multi}. All entities appearing in the abstract are included in our keyphrase bank. 
The state-of-the-art system~\cite{koncel-kedziorski-etal-2019-text} exploits the scientific entities, their relations, and the relation types.
In our setup, we ignore the relation graph, and focus on generating the abstract with only entities and title as the input.
Due to the dataset's relatively uniform language style and smaller size, we do not experiment with our style specification component. 

%% file: experiment.tex
\subsection{Implementation Details}
For argument generation, we truncate the input OP and retrieved passages to $500$ and $400$ words. Passages are optionally appended to OP as our encoder input. 
The keyphrase bank size is limited to $70$ for argument, and $30$ for Wikipedia and AGENDA data (based on the average numbers in Table~\ref{tab:data-stat}), with keyphrases truncated to $10$ words. We use a vocabulary size of $50$K for all tasks.

\smallskip
\noindent \textbf{Training Details.} 
Our models use a two-layer LSTM for both decoders. They all have $512$-dimensional hidden states per layer and dropout probabilities~\cite{NIPS2016_6241} of $0.2$ between layers.
Wikipedia titles are encoded with the summation of word embeddings due to their short length. 
The learning process is driven by AdaGrad~\cite{duchi2011adaptive} with $0.15$ as the learning rate and $0.1$ as the initial accumulator. We clip the gradient norm to a maximum of $2.0$. The mini-batch size is set to $64$. And the optimal weights are chosen based on the validation loss.

For argument generation, we also pre-train the encoder and the lower layer of realization decoder using language model losses. 
We collect all the OP posts from the training set, and an extended set of reply paragraphs, which includes additional counter-arguments that have non-negative \texttt{karma}.
For Wikipedia, we consider the large collection of $1.9$ million unpaired normal English Wikipedia paragraphs to pre-train the model for both normal and simple Wikipedia generation. 

\noindent \textbf{Beam Search Decoding.} 
For inference, we utilize beam search with a beam size of $5$. We disallow the repetition of trigrams, and replace the $\texttt{UNK}$ with the keyphrase of the highest attention score.

\subsection{Baselines and Comparisons}
For all three tasks, we consider a \textsc{Seq2seq} with attention baseline~\cite{DBLP:journals/corr/BahdanauCB14}, which encodes the input text and keyphrase bank as a sequence of tokens, and generates the output. 

For argument generation, we implement a \textsc{Retrieval} baseline, which returns the highest reranked passage retrieved with OP as the query. 
We also compare with our prior model~\cite{hua-wang-2018-neural}, which is a multi-task learning framework to generate both keyphrases and arguments.

For Wikipedia generation, a \textsc{Retrieval} baseline obtains the most similar paragraph from the training set with input title and keyphrases as the query, measured with bigram cosine similarity. We further train a logistic regression model (\textsc{LogRegSel}), which takes the summation of word embeddings in a phrase and predicts its inclusion in the output for a normal or simple Wiki paragraph.

For abstract generation, we compare with the state-of-the-art system \textsc{GraphWriter}~\cite{koncel-kedziorski-etal-2019-text}, which is a transformer model enabled with knowledge graph encoding mechanism to handle both the entities and their structural relations from the input. 

We also report results by our model variants to demonstrate the usefulness of content planning and style control: (1) with gold-standard\footnote{``Gold-standard" indicates the keyphrases {that have content word overlap with the reference sentence}.} 
keyphrase selection for each sentence ({Oracle Plan.}), and (2) without style specification.

%% file: analysis.tex
\subsection{Automatic Evaluation}
We report precesion-oriented BLEU~\cite{papineni-etal-2002-bleu}, recall-oriented ROUGE-L~\cite{W04-1013} that measures the longest common subsequence, and METEOR~\cite{denkowski-lavie:2014:W14-33}, which considers both precision and recall.

\noindent \textbf{Argument Generation.} 
For each input OP, there can be multiple possible counter-arguments. We thus consider the best matched (i.e., highest scored) reference when reporting results in Table~\ref{tab:arggen-results}. 
Our models yield significantly higher BLEU and ROUGE scores than all comparisons while producing longer arguments than generation-based approaches. Furthermore, among our model variants, oracle content planning further improves the performance, indicating the importance of content selection and ordering. Taking out style specification decreases scores, indicating the influence of style control on generation.\footnote{We do not compare with our recent model in \newcite{hua-etal-2019-argument-generation} due to the training data difference caused by our new sentence style scheme. However, the newly proposed model generates arguments with lengths closer to human arguments, benefiting from the improved content planning module.}

\noindent \textbf{Wikipedia Generation.} 
Results on Wikipedia (Table~\ref{tab:wikigen-results}) show similar trends, where our models almost always outperform all comparisons across metrics. 
The significant performance drop on ablated models without style prediction proves the effectiveness of style usage. Our model, if guided with oracle keyphrase selection per sentence, again achieves the best performance. 

\begin{figure}[t]
\subfloat
{
    {\includegraphics[width=37mm]{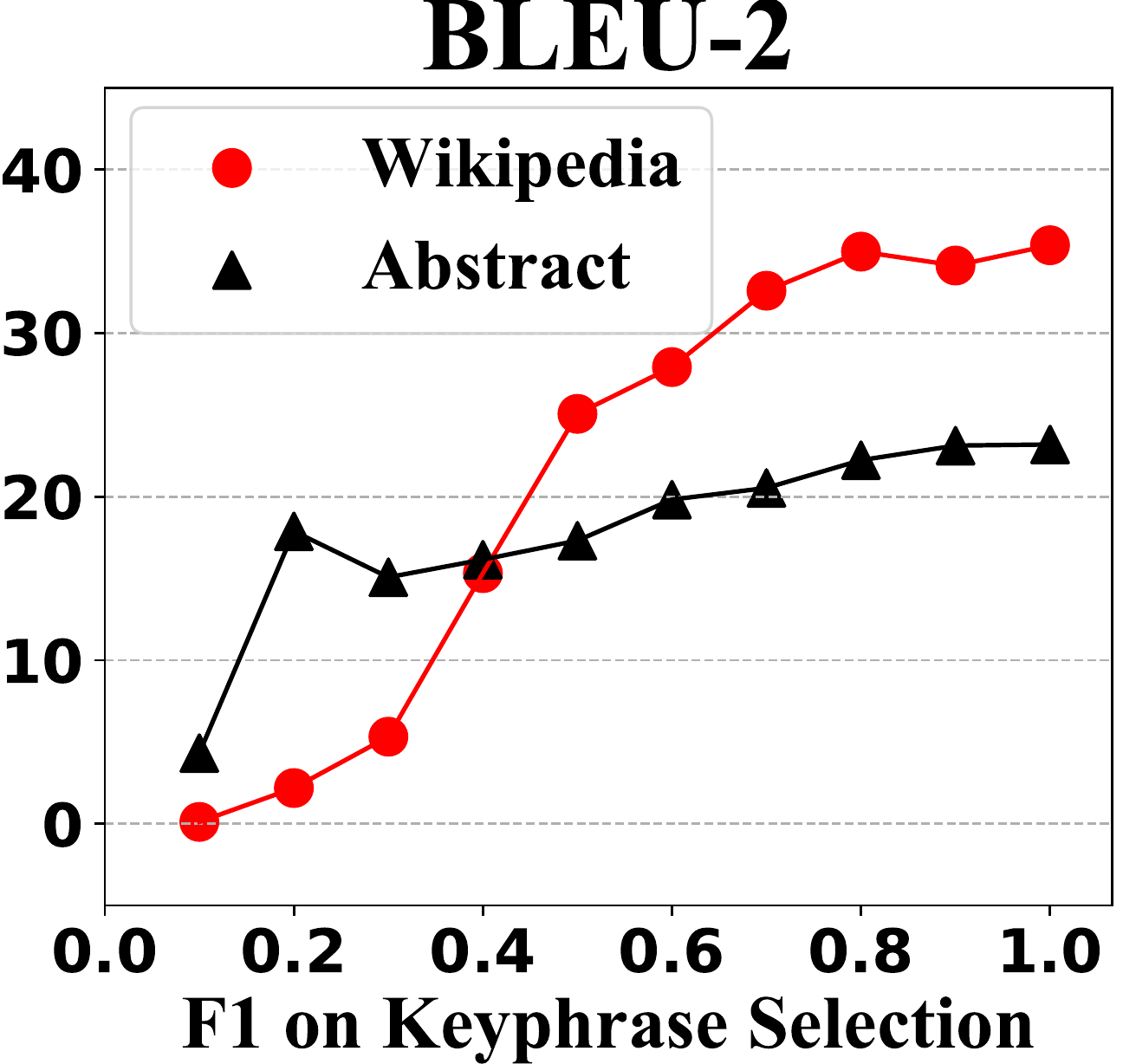}}
}
\subfloat
{
\includegraphics[width=37mm]{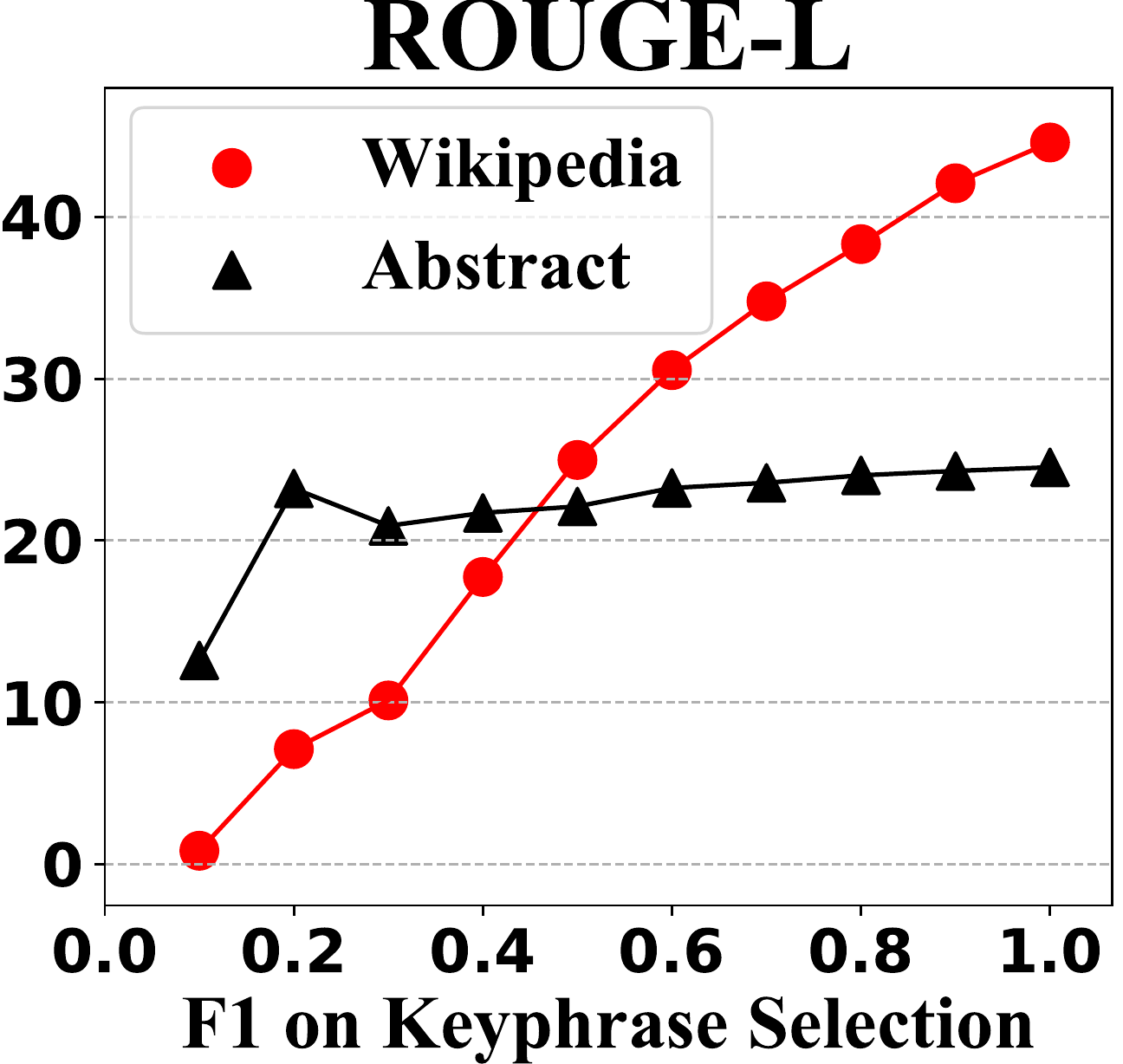}
}
\vspace{-2mm} 
\caption{
Effect of keyphrase selection (F1 score) on generation performance, measured by BLEU and ROUGE. Positive correlations are observed.
}
\label{fig:gen-by-plan}
\end{figure}

We further show the \textit{effect of content selection on generation} on Wikipedia and abstract data in Figure~\ref{fig:gen-by-plan}, where we group the test samples into $10$ bins based on F1 scores on keyphrase selection.\footnote{We calculate F1 by aggregating the selections across all sentences. For argument generation, keyphrases are often paraphrased, making it difficult to calculate F1 reliably, therefore omitted here.}  
We observe a strong correlation between keyphrase selection and generation performance, e.g., for BLEU, Pearson correlations of $0.95$ ($p<10^{-4}$) and $0.85$ ($p<10^{-2}$) are established for Wikipedia and abstract. For ROUGE, the values are $0.99$ ($p<10^{-8}$) and $0.72$ ($p<10^{-1}$).

\begin{table}[t]
\fontsize{10}{11}\selectfont
 \setlength{\tabcolsep}{0.7mm}
   \centering
\centering
    \begin{tabular}{lllll}
    \toprule
    & \textbf{BLEU} & \textbf{ROUGE} & \textbf{MTR} & \textbf{Len.} \\
    \midrule
    \textsc{Retrieval} & 7.81 & 15.68 & {\bf 10.59} & 150.0 \\
    \textsc{Seq2seq} & 3.64 & 19.00 & 9.85 & 51.7\\
    H\&W~\shortcite{hua-wang-2018-neural} & 5.73 & 14.44 & 3.82 & 36.5 \\
    \midrule
    \textsc{Ours} (Oracle Plan.) & 16.30$^\ast$ & 20.25$^\ast$ & 11.61 & 65.5 \\
    \textsc{Ours} & {\bf 13.19}$^\ast$ & 20.15$^\ast$ & 10.42 & 65.2\\
    \quad w/o Style & 12.61$^\ast$ & {\bf 20.28}$^\ast$ & 10.15 & 64.5 \\
    \quad w/o Passage & 11.84$^\ast$ & 19.90$^\ast$ & 9.03 & 62.6 \\

    \bottomrule
    \end{tabular}
    \caption{
    Results on argument generation with BLEU (up to bigrams), ROUGE-L, and METEOR (MTR). 
    Best systems without oracle planning are in {\bf bold} per metric. Our models that are significantly better than all comparisons are marked with $\ast$ ($p<0.001$, approximate randomization test~\cite{noreen1989computer}).
    \label{tab:arggen-results}}
    \vspace{-3mm}
\end{table}

\begin{table*}[t]
\fontsize{10}{11}\selectfont
 \setlength{\tabcolsep}{1.2mm}
   \centering
    \begin{tabular}{lllllllll}
    \toprule
    & \textbf{BLEU} & \textbf{ROUGE} & \textbf{METEOR} & \textbf{Length} & \textbf{BLEU} & \textbf{ROUGE} & \textbf{METEOR} & \textbf{Length} \\
    \midrule
    & \multicolumn{4}{c}{\bf{Normal Wikipedia}} & \multicolumn{4}{c}{\bf{Simple Wikipedia}} \\
    \cmidrule{2-5}  \cmidrule{6-9} 
    \textsc{Retrieval} & 20.10 & 28.60 & 12.23 & 44.5 & 21.99 & 33.44 & 12.97 & 34.7 \\
    \textsc{Seq2seq} & 22.62 & 27.49 & 14.74  & 52.9 & 21.98  & 29.36  & 16.94  & 52.8 \\
     \textsc{LogRegSel} & 29.28 & 28.65 & {\bf 27.76} & 34.3 & 5.59 & 23.21 & 13.27 & 13.0  \\
    \hdashline
    \textsc{Ours} (Oracle Plan.) & 37.70$^\ast$ & 45.41$^\ast$ &  31.65$^\ast$ & 79.8 & 34.22$^\ast$ & 45.48$^\ast$ & 32.84$^\ast$ & 70.5 \\
    \textsc{Ours} & {\bf 33.76}$^\ast$ & {\bf 40.08}$^\ast$ & 25.70 & 65.4 & {\bf 31.22}$^\ast$ & {\bf 40.76}$^\ast$ & {\bf 26.76}$^\ast$ & 58.7  \\
    \quad w/o Style & 31.06$^\ast$ & 37.72$^\ast$ & 24.56 & 71.0 & 27.94$^\ast$  & 38.20$^\ast$ & 25.87$^\ast$ & 64.5 \\
    \bottomrule
    \end{tabular}
    \caption{Results on Wikipedia generation. Best results without oracle planning are in {\bf bold}. 
    $\ast$: Our models that are significantly better than all comparisons ($p<0.001$, approximate randomization test).
    \label{tab:wikigen-results}}
\end{table*}

\begin{table}[t]
\fontsize{10}{11}\selectfont
 \setlength{\tabcolsep}{0.7mm}
   \centering
\centering
    \begin{tabular}{lllll}
    \toprule
    & \textbf{BLEU} & \textbf{ROUGE} & \textbf{MTR} & \textbf{Len.} \\
    \midrule
    \textsc{GraphWriter} & 29.95 & 28.56 & 19.90 & 130.1 \\
    \textsc{Seq2seq} & 18.13 & 21.03 & 13.95 & 134.8 \\
    \midrule
    \textsc{Ours} (Oracle Plan.) & 25.03 & 26.18 & 19.21 & 125.8 \\
    \textsc{Ours} & 20.32 & 23.30 & 15.95 & 128.3 \\
    \bottomrule
    \end{tabular}
    \caption{Results on paper abstract generation. Notice that \textsc{GraphWriter} models rich information about relations and relation types among entities, which is not utilized by our model. 
    } 
    \label{tab:absgen-results}
    \vspace{-2mm}
\end{table}

\noindent \textbf{Abstract Generation.}
Lastly, we compare with the state-of-the-art \textsc{GraphWriter} model on AGENDA dataset in Table \ref{tab:absgen-results}. 
Although our model does not make use of the relational graph encoding, we achieve competitive ROUGE-L and METEOR scores given the oracle plans. 
Our model also outperforms the seq2seq baseline, which has the same input, indicating the applicability of our method across different domains. 

\subsection{Human Evaluation}

\begin{table}[t]
\fontsize{9.5}{11}\selectfont
 \setlength{\tabcolsep}{0.4mm}
    \centering
    \begin{tabular}{lccccccc}
    \toprule
    & \multicolumn{3}{c}{\textbf{Argument}} &  \phantom{} & \multicolumn{3}{c}{\textbf{Wikipedia}} \\
    \cmidrule{2-4} \cmidrule{6-8} 
    &  \textbf{Gram.} & \textbf{Corr.} & \textbf{Cont.}  &\phantom{} & \textbf{Gram.} & \textbf{Corr.} & \textbf{Cont.}\\
         \midrule
        \textsc{Human}  &  4.81 & 3.90 & 3.48 & \phantom{} & 4.84 & 4.73 & 4.49 \\
        \textsc{Ours} &  \bf{3.99}$^\ast$ & \bf{2.78}$^\ast$ & \bf{2.61}$^\ast$  & \phantom{} & \bf{3.38} & \bf{3.24}$^\ast$ & 3.43 \\
        \quad w/o Style & 3.03 & 2.26 & 2.03 & \phantom{} & 2.99 & 2.89 & {\bf 3.50}   \\ 
        \hdashline
        {\it Krippendorff's} $\alpha$ & 0.75 & 0.69 & 0.33  & \phantom{} & 0.70 & 0.56 & 0.55   \\
        \bottomrule
    \end{tabular}
    \caption{
    Human evaluation on argument generation (Upper) and Wikipedia generation (Bottom). 
    Grammaticality ({\bf Gram}), correctness ({\bf Corr}), and content richness ({\bf Cont}) are rated on Likert scale ($1-5$). 
    We mark our model with $\ast$ to indicate statistically significantly better ratings over the variant without style specification ($p<0.001$, approximate randomization test).
    }
    \label{tab:human-eval}
    \vspace{-6mm}
\end{table}

We further ask three proficient English speakers to assess the quality of generated arguments and Wikipedia paragraphs. Human subjects are asked to rate on a scale of $1$ (worst) to $5$ (best) on \textbf{grammaticality}, \textbf{correctness} of the text (for arguments, the stance is also considered), and \textbf{content richness} (i.e., coverage of relevant points). Detailed guidelines for different ratings are provided to the raters (see Supplementary). 
For both tasks, we randomly choose $30$ samples from the test set; outputs from two variants of our models and a human written text are presented in random order.  

According to Krippendorff's $\alpha$, the raters achieve substantial agreement on grammaticality and correctness, while the agreement on content richness is only moderate due to its subjectivity.
As shown in Table~\ref{tab:human-eval}, on both tasks, our models with style specification produce more fluent and correct generations, compared to the ones without such information. However, there is still a gap between system generations and human edited text.

We further show {\bf sample outputs}  in Figure~\ref{fig:sample-outputs}. 
The first example is on the topic of abortion, our model captures the relevant concepts such as ``\textit{fetuses are not fully developed}'' and ``\textit{illegal to kill}''.  
It also contains fewer repetitions than the seq2seq baseline. 
For Wikipedia, our model is not only better at controlling the global simplicity style, but also more grammatical and coherent than the seq2seq output.

\subsection{Further Analysis and Discussions}

We further investigate the usage of different styles, and show the top frequent patterns for each argument style from human arguments and our system generation (Table~\ref{tab:style-samples}).
We first calculate the most frequent $4$-grams per style, then extend it with context. We manually cluster and show the representative ones. 
For both columns, the popular patterns reflect the corresponding discourse functions: \textsc{Claim} is more evaluative, \textsc{Premise} lists out details, and \textsc{Functional} exhibits argumentative stylistic languages. 
{Interestingly, our model also learns to paraphrase popular patterns, e.g., ``\textit{have the freedom to}'' vs. ``\textit{have the right to}''.} 

For Wikipedia, the sentence style is defined by length. To validate its effect on content selection, we calculate the average number of keyphrases per style type. The results on human written paragraphs are $2.0$, $3.8$, $5.8$, and $9.0$ from the simplest to the most complex. A similar trend is observed in our model outputs, which indicates the challenge of content selection in longer sentences. 

For future work, improvements are needed in both model design and evaluation. 
As shown in Figure~\ref{fig:sample-outputs}, system arguments appear to overfit on stylistic languages and rarely create novel concepts like humans do.
Future work can lead to improved model guidance and training methods, such as reinforcement learning-based explorations, and better evaluation to capture diversity.

\begin{table}[t]
\fontsize{9}{10}\selectfont
\setlength{\tabcolsep}{0.6mm}
    \centering
    \begin{tabular}{|c|p{35mm}|p{35mm}|}
        \hline
        & \textbf{Human} & \textbf{Our model}  \\
        \hline
        \textsc{C} & It doesn't mean that; 
        everyone should be able to 
        & 
        I don't believe that it is necessary; 
        don't need to be able to\\
        \hline
        \textsc{P} & have the freedom to; 
        is leagal in the US; imagine for a moment if; 
        Let's say (you/your partner/a friend) 
        & 
        have the right to (bear arms/cast a ballot vote); 
        For example, (if you look at/let's look at/I don't think)  \\
        \hline
        \textsc{F} & Why is that?; 
        that's ok; 
        Would it change your mind?  
        & 
        I'm not sure (why/if) this is; 
        TLDR: I don't care about this\\
        \hline
    \end{tabular}
    \caption{Top frequent patterns captured in style \textsc{Claim} (\textsc{C}), \textsc{Premise} (\textsc{P}), and \textsc{Functional} (\textsc{F}) from arguments by human and our model. }
    \label{tab:style-samples}
\end{table}

%% file: conclusion.tex
We present an end-to-end trained neural text generation model that considers sentence-level content planning and style specification to gain better control of text generation quality. Our content planner first identifies salient keyphrases and a proper language style for each sentence, then the realization decoder produces fluent text. 
We consider three tasks of different domains on persuasive argument generation, paragraph generation for normal and simple versions of Wikipedia, and abstract generation for scientific papers. 
Experimental results demonstrate the effectiveness of our model, where it obtains significantly better BLEU, ROUGE, and METEOR scores than non-trivial comparisons. Human subjects also rate our model generations as more grammatical and correct when language style is considered.

\begin{figure}[t]
	\fontsize{8.5}{10}\selectfont
	\setlength{\tabcolsep}{0.8mm}
	\begin{tabular}{|p{75mm}|}
	\hline
	\textbf{Topic}: Aborting a fetus has some non-zero negative moral implications \\
    \hline 
    
    \textbf{Human}: It's not the birthing process that changes things. It's the existence of the baby. Before birth, the baby only exists inside another human being. After birth, it exists on its own in the world like every other person in the world. 
    \\
    
	\textbf{Seq2seq}: i 'm not going to try to change your view here , but i do n't want to change your position . i do n't think it 's fair to say that a fetus is not a person . it 's not a matter of consciousness . \\
    \rowcolor{lightgray!30}
    \textbf{Our model}: tl ; dr : i agree with you , but i think it 's important to note that fetuses are not fully developed . i do n't know if this is the case , but it does n't seem to be a compelling argument to me at all , so i 'm not going to try to change your view by saying that it should be illegal to kill\\
    \hline
	\end{tabular}
	
	\begin{tabular}{|p{75mm}|}
	\hline
	\textbf{Topic}: Moon Jae-in \\
	\hline
	
	\textbf{Simple Wikipedia}: Moon Jae-in is a South Korean politician. He is the 12th and current President of South Korea since 10 May 2017 after winning the majority vote in the 2017 presidential election. \\
	\textbf{Seq2seq:} moon election park is a election politician who served as prime minister of korea from 2007 to 2013 . he was elected as a member of the house of democratic party in the moon 's the the moon the first serving president of jae-in , in office since 2010 . \\
	\rowcolor{lightgray!30}
	\textbf{Our model}: moon jae-in is a south korean politician and current president of south korea from 2012 to 2017 and again from 2014 to 2017. \\
	\hline
	
	\textbf{Normal Wikipedia}: Moon Jae-in is a South Korean politician serving as the 19th and current President of South Korea since 2017. He was elected after the impeachment of Park Geun-hye as the candidate of the Democratic Party of Korea. \\
	
	\textbf{Seq2seq}: moon winning current is a current politician who served as prime minister of korea from 2007 to 2013 . he was elected as a member of the house of democratic party in the moon 's the the current the first president of pakistan , in office . prior to that , he also served on the democratic republic of germany .\\
	\rowcolor{lightgray!30}
	\textbf{Our model}:  moon jae-in is a south korean politician serving as the 19th and current president of south korea , since 2019 to 2019 and 2019 to 2017 respectively he has been its current president ever since . \\
	\hline
	\end{tabular}
	\vspace{-2mm}
	\caption{
	Sample outputs for argument generation and Wikipedia generation. 
   }
\label{fig:sample-outputs}
\vspace{-2mm}
\end{figure}

%% file: appendix.tex
\subsection{Rule-based Argumentative Style Label Construction.}
In \S 4, we mention a set of rules to automatically label sentence styles for argument generation. The goal is to capture the argumentative discourse function using common patterns. The complete version of the rules for \textsc{Claim} and \textsc{Premise} are listed in Table \ref{tab:style-rules}.

\subsection{Wikipedia Sentence Length Distribution.}
\begin{figure}[h]
\centering
  \includegraphics[height=30mm]{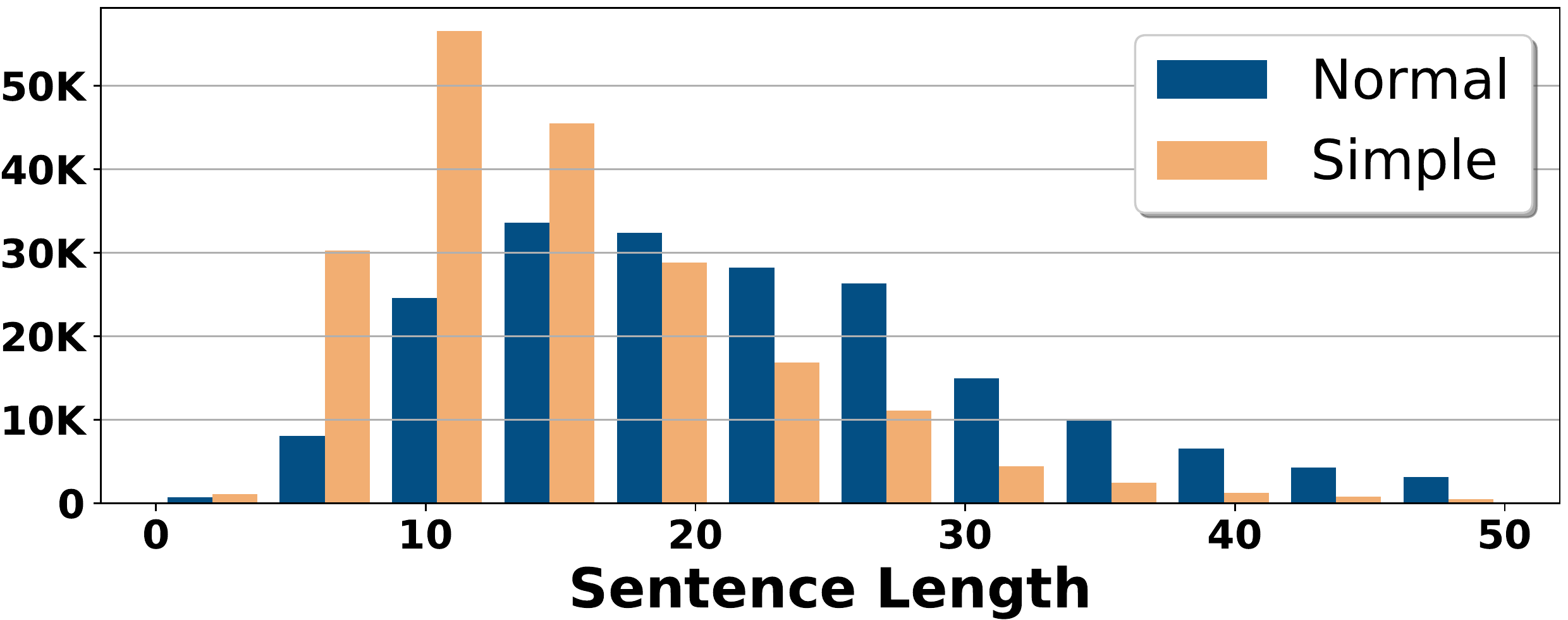}
  \caption{
  The distribution of sentence length in normal and simple Wikipedia. The normal Wikipedia contains more sentences in the longer range, and the opposite is true for its simple counterpart.
  }
  \label{fig:style_dist}
\end{figure}

As is described in \S 4.2, we assign the sentence style labels for Wikipedia data based on the sentence length. 
In Figure \ref{fig:style_dist} we show the distribution of sentence lengths for normal and simple Wikipedia. For both versions, the majority fall into the $[10, 30]$ range. The normal version tends to contain more longer sentences and fewer shorter sentences than its simple counterpart. Based on these observations, we choose the four intervals with a step size of $10$ words for length categories, to ensure the balance of data samples across categories.

\subsection{Human Evaluation Guidelines}
We conduct human evaluation on the generated arguments and Wikipedia paragraphs. 
For argument generation, we randomly choose $33$ topics from the test set, from which the first $3$ topics are used merely for the human judges to calibrate among their own standards. The remaining $30$ topics are used for the final evaluation.
We show the guidelines for evaluation on argument data in Table \ref{tab:arggen-human}.

On Wikipedia data, we randomly select $16$ topics from the test set, and present both the normal and simple output during the evaluation. The first topic ($2$ samples) are used for rater calibration, while the rest $15$ topics ($30$ samples) are kept for analysis. The guidelines are listed in Table \ref{tab:wikigen-human}.

For both tasks, we consider our model and an ablated model where style specification is disabled. We present these two system outputs alongside human constructed ones, and shuffle them for each sample to eliminate the biases associated with the order.

\subsection{Sample Output}
We show more sample outputs on all three tasks from Figure \ref{tab:arggen-sample-1} to \ref{tab:absgen-sample-2}. We highlight our model generation among the human constructed texts and oracle plan guided generation.

%% file: tables.tex
\begin{table*}[t]
\fontsize{9}{11}\selectfont
\centering
    \begin{tabular}{|lp{130mm}|}
    \hline
    \textbf{Rule} & \textbf{Patterns} \\
    \hline
    \multicolumn{2}{|c|}{\textsc{Claim}} \\
    \hline
    Belief  & \texttt{i (don't)? (believe|agree|concede|suspect|doubt|see|feel|understand)}  \\
    \hline
    Imperative & \texttt{(any|anyone|anybody|every|everyone|everybody|all|most|few|no|no one} \\
    & \texttt{|nobody|it|we|you|they|there) \textbackslash w\{0,10\} (could|should|might|need|must)}\\
    \hline
    Sense & \texttt{(it|this|that) make (no|zero)? sense} \\
    \hline
    Chance & \texttt{(chance|likelihood|possibility|probability) .}$^\ast$ \texttt{(slim|zero|negligible)} \\
    \hline
    Evaluation & \texttt{(be|seem) (necessary|unnecessary|moral|immoral|right|wrong|stupid}\\
    & \texttt{|unconstitutional|costly|inefficient|efficient|reasonable|beneficial}\\
    & \texttt{|important|unfair|harmful|justified|jeopardized|meaningless|flawed} \\
    & \texttt{|justifiable|unacceptable|impossible|irrational|foolish)} \\
    \hline
    Miscellaneous & \texttt{(in my opinion|imo|my view|i be try to say|have nothing to do with|tldr)} \\
    \hline 
    \multicolumn{2}{|c|}{\textsc{Premise}} \\
    \hline
    Affect & \texttt{(help|improve|reduce|deter|increase|decrease|promote)} \\
    \hline
    Example & \texttt{(for example|for instance|e.g.)}\\
    \hline
    \end{tabular}
    \caption{Patterns for sentence style label construction on \textsc{Claim} and \textsc{Premise} for argument generation.}
    \label{tab:style-rules}
\end{table*}

\begin{table*}[b]
    \centering\fontsize{9}{11}\selectfont
    \begin{tabular}{|p{135mm}|}
         \hline
         In the following survey, you will read 33 short argumentative text prompts and evaluate 3 counter-arguments for each of them. Please
        rate each counter-argument on a scale of 1-5 (the higher the better), based 
        on the following three aspects:  
        {\fontsize{9}{11}\selectfont
        \begin{itemize}[leftmargin=3mm]
            \item {\bf Grammaticality:} whether the counter-argument is fluent and has no grammar
            errors
            {\fontsize{9}{11}\selectfont
            \begin{itemize}[leftmargin=3mm]
                \item 1. \textit{the way the way etc. 'm not 's important}
                \item 3. \textit{is a good example. i don't think should be the case. i're not going to talk whether or not it's bad.}
                \item 5. \textit{i agree that the problem lies in the fact that too many representatives do n't understand the issues or have money influencing their decisions.}
            \end{itemize}}
            \item {\bf Correctness:}  whether the counter-argument is relevant to the topic and of correct stance
            \begin{itemize}[leftmargin=3mm]
                \item 1. \textit{i don't think it 's fair to say that people should n't be able to care for their children}
                \item 3. \textit{i don't agree with you and i think legislative bodies do need to explain why they vote that way}
                \item 5. \textit{there are hundreds of votes a year . how do you decide which ones are worth explaining ? so many votes are bipartisan if not nearly unanimous . do those all need explanations ? they only have two years right now and i do n't want them spending less time legislating .}
            \end{itemize}
            \item {\bf Content richness:} whether the counter-argument covers many talking points
            \begin{itemize}[leftmargin=3mm]
                \item 1. \textit{i do n't agree with your point about legislation but i 'm not going to change your view.}
                \item 3. \textit{i agree that this is a problem for congress term because currently it is too short.}
                \item 5. \textit{congressional terms are too short and us house reps have to spend half of their time campaigning and securing campaign funds. they really have like a year worth of time to do policy and another year to meet with donors and do favors.}\vspace{-5mm}
            \end{itemize}
        \end{itemize}} 
         \\
         \hline
    \end{tabular}
    \caption{Evaluation guidelines on argument data and representative examples on rating scales.}
    \label{tab:arggen-human}
\end{table*}

\begin{table*}[h]
    \centering\fontsize{9}{11}\selectfont
    \begin{tabular}{|p{135mm}|}
         \hline
         In the following survey, you will read 32 samples. Each of them contains a topic, and 3 text snippets explaining the topic. For each of the explanation, please rate it on a scale of 1-5 (the higher the better) based on the following three aspects:\\
        {\fontsize{9}{11}\selectfont
        \begin{itemize}[leftmargin=3mm]
            \item {\bf Grammaticality:} whether the explanation is fluent and has no grammar errors 
            \begin{itemize}[leftmargin=3mm]
                \item 1. \textit{android android operating system mobile kindle is a modified operating}
                \item 3. \textit{android is an operating system for mobile mobile mobile and other manufactures like htc and the}
                \item 5. \textit{android is an operating system for mobile devices . it is also used by other manufactures like htc and samsung .
                }
            \end{itemize}
            \item {\bf Correctness:} whether the explanation contains obvious semantic mistakes or contradictions. Please note that this is NOT intended for fact checking, so you should not find other resources to determine if the concrete information (such as years, locations) are wrong, instead please apply commonsense level knowledge to judge the correctness
            \begin{itemize}[leftmargin=3mm]
                \item 1. \textit{android is used for tablets such as amazon.com as well as other phone such as linux and amazon}
                \item 3. \textit{android is an operating system for android and devices .}
                \item 5. \textit{android is an operating system for mobile devices .}
            \end{itemize}
            \item {\bf Content richness:} whether the explanation covers the amount of information that is necessary to explain the topic
            \begin{itemize}[leftmargin=3mm]
                \item 1. \textit{modified mobile mobile android}
                \item 3. \textit{android is an operating system used for mobile devices .}
                \item 5. \textit{android is an operating system for mobile devices , it is mostly used for, like google 's own google pixel, as well as by other phone manufacturers like htc and samsung .}\vspace{-5mm}
            \end{itemize}
        \end{itemize}} 
        \\
         \hline
    \end{tabular}
    \caption{Evaluation guidelines on Wikipedia data and representative examples on rating scales.}
    \label{tab:wikigen-human}
\end{table*}

\clearpage

\begin{figure*}[t]
    \centering
	\begin{tabular}{|p{140mm}|}
    \hline
	\textbf{Topic}: CMV: All alleged crimes or infractions involving a police officer should be investigated by a higher-level agency. \\
	\hline
	\textbf{Human: }  if it was required to take this seriously , people who want to protest the police would just file complaints against all the police in a department constantly . are you going to make it a crime to file a complaint if the review board determines there is n't enough evidence to convict the cop ? they could just keep the complaint vague to avoid charges of filing false reports . every person who has a cop car pass them could file a complaint that the cop is stalking them. \\
	\rowcolor{lightgray!30}
	\textbf{Our model:} jail sentences for example are a result of negligence . if you do n't like it , you 're going to have to deal with a lot of bad lawyers , and you 'll have to prove you 're guilty of a crime in the first place and you 're not going to be able to sue for damages if you 're in jail , you should n't have to\\
	\textbf{Our model (Oracle Plan):} additionally , there is a difference between the two cases in the united states and between the federal government and the department of justice . in the case of a police officer , the fbi has a duty to report a crime , and it is not required to report the crime . this is not the case , it is the victim 's job to determine whether or not\\
	\hline
	\end{tabular}
	\caption{
	Sample outputs on argument generation.
  }\label{tab:arggen-sample-1}
\end{figure*}

\begin{figure*}[th]
    \centering
	\begin{tabular}{|p{140mm}|}
    \hline
	\textbf{Topic}: CMV:High Heeled shoes are just as "oppressive" as the face veil/niqab \\
	\hline
	\textbf{Human: }  the niqab has the unique property of covering a person 's face , which is where we tend to ascribe people 's personality and emotion . in any media critique , calling someone faceless means that they lack humanization . it 's usually the default way to mark someone as either a villain or an acceptable target for consequence-free violence .\\
	\rowcolor{lightgray!30}
	\textbf{Our model:} i live in a small town in the united states . there are a lot of things i do n't want to do , but it 's not a bad thing for me to think about it as a matter of personal experience , and i think it 's important to keep in mind that it 's something that can be seen as a good thing in the long run \\
	\textbf{Our model (Oracle Plan):} lastly , a woman 's right to bodily autonomy is not the same as being a woman . it 's not a matter of whether or not a woman has a right to her body , it 's about her ability to make decisions about her own body . she has the right to do whatever she wants with her body and her body is her right to use her \\
	\hline
	\end{tabular}
	\caption{
	Sample outputs on argument generation.
  }\label{tab:arggen-sample-2}
\end{figure*}


\begin{figure*}[th]
    \centering
	\begin{tabular}{|p{140mm}|}
    \hline
	\textbf{Topic}: Breaking Bad \\
	\hline
	\textbf{Simple Wikipedia: }
	Breaking Bad is an American television series set in Albuquerque, New Mexico. It started in January 2008. The show was broadcast across Canada and the United States on cable channel AMC. It has won 10 Emmy Awards. Breaking Bad ended in September 2013. Bryan Cranston plays the main role, Walter White. There are five seasons and 62 episodes. It is about a chemistry teacher who is told he has lung cancer and starts making the illegal drug methamphetamine to pay for his family's needs after he dies. Breaking Bad was made by Vince Gilligan. \\
	\rowcolor{lightgray!30}
	\textbf{Our model:} bad breaking is an american television series set in albuquerque , new mexico . it started in january 2008 and on january 20 , 2011 .\\
	\textbf{Our model (Oracle Plan):} breaking bad is an american television series set in albuquerque , new mexico . it started in january 2008 . the show has been broadcast across canada and the united states on cable channel amc . it has won emmy awards 10 and 3 respectively since it ended in september 2013 ended on september 2013 after breaking bad breaking itself in 2013 and 2013 respectively cable channel channel 2 ended in 2013 ) cable cable channel cable 10 . \\
	\hline
	\textbf{Normal Wikipedia: } Breaking Bad is an American neo-Western crime drama television series created and produced by Vince Gilligan. The show originally aired on AMC for five seasons, from January 20, 2008 to September 29, 2013. Set and filmed in Albuquerque, New Mexico, the series tells the story of Walter White, a struggling and depressed high school chemistry teacher who is diagnosed with stage-3 lung cancer. Together with his former student Jesse Pinkman, White turns to a life of crime by producing and selling crystallized methamphetamine to secure his family's financial future before he dies, while navigating the dangers of the criminal world. The title comes from the Southern colloquialism "breaking bad" which means to "raise hell" or turn to a life of crime. \\
	\rowcolor{lightgray!30}
	\textbf{Our model:}  bad breaking is an american an american neo-western crime drama television series an american neo-western crime drama television series drama series television written and produced by vince vince . the show aired on amc for five seasons from january 20 , 2011 to september 29 , 2013 . \\
	\textbf{Our model (Oracle Plan):} bad breaking is an american an american neo-western crime drama television series an american neo-western crime drama television series drama series television written and produced by vince gilligan . the show originally aired on amc for five seasons from january 20 , 2011 to september 29 , 2013 . the series was filmed in filmed in albuquerque , new mexico and tells the story of walter walter , a highly high school depressed depressed chemistry school diagnosed with lung with lung cancer . the novel follows a life of student and former jesse SOS student , walter white methamphetamine , and navigating the navigating of his family in the world the the world his future financial and depressed high school chemistry teacher ' depressed high school chemistry teacher SOS ' `` the breaking bad SOS " is about a life in the criminal world , with white and walter white methamphetamine and the navigating the dangers of financial financial and depressed high school in new mexico 's financial financial , an american neo-western crime drama television series .luca luca badoer was an formula one italian racing driver . he was born in born in montebelluna . he raced for the scuderia the minardi , forti corse corse , forti corse and teams ferrari and raced for his raced for italia raced for raced for the scuderia italia raced for them as well as for raced for the scuderia italia for the raced for both the raced for the scuderia italia in addition to his racing duties , he also raced for and raced internationally for the prix de forti corse and competed for the for the right and raced against the scuderia raced for racing with the for for his racing he was also active in the european prix for reserve reserve racing , qualifying for the grand prix de corse and michael michael 's ferrari teams as well .
\\
	\hline
	\end{tabular}
	\caption{
	Sample outputs on Wikipedia generation.
  }\label{tab:wikigen-sample-1}
\end{figure*}

\begin{figure*}[th]
    \centering
	\begin{tabular}{|p{140mm}|}
    \hline
	\textbf{Topic}: Kingdom of Iceland \\
	\hline
	\textbf{Simple Wikipedia: }
	The Kingdom of Iceland was a constitutional monarchy that had a personal union with Denmark. The kingdom was created on 1 December 1918. It lasted until 17 June 1944 when a national referendum created the Republic of Iceland. \\
	\rowcolor{lightgray!30}
	\textbf{Our model:} the kingdom of iceland was a monarchy in the kingdom and union of iceland . it was created on 1 december created out of a personal union , the republic of denmark , in 1918 – 1945 .\\
	\textbf{Our model (Oracle Plan):} the kingdom of iceland was a personal union of iceland and union with a monarchy . the kingdom was created on 1 december 1 , 1944 . it lasted until 1944 , when iceland officially created the republic of iceland in a national referendum . iceland created the created republic of denmark , which lasted until june until a national national referendum in 1944 . the name `` iceland " lasted until 1945 , when it became a a a national a referendum . \\
	\hline
	\textbf{Normal Wikipedia: }The Kingdom of Iceland was a constitutional monarchy, a sovereign and independent country that was established by the Act of Union with Denmark signed on 1 December 1918. It lasted until 17 June 1944 when a national referendum established the Republic of Iceland in its place. \\
	\rowcolor{lightgray!30}
	\textbf{Our model:}  the kingdom of iceland , officially the kingdom `` iceland " , was a constitutionally independent country and a sovereign of the kingdom . it was established by the act of 1918 and signed on 1 december 1 , 1918 , in iceland , union of denmark , and denmark and denmark , where it existed until it ceased to exist until it lasted until june 17 , 1945 , iceland was officially established under the act on 1 may 1949 . \\
	\textbf{Our model (Oracle Plan):} the kingdom of iceland was a monarchy established by the act of union of a sovereign and independent country signed on 1 december 1 , 1918 . it lasted until 1944 , when iceland established the republic as a national referendum and took place in 1944 . the name `` iceland " lasted until 12 june june , when it established the established republic of iceland . the `` a national referendum " took place after a a a national national national a constitution that established the the republic that lasted until june 1944 . it existed until 2 june until a national a referendum took place outside iceland in may 1944 .\\
	\hline
	\end{tabular}
	\caption{
	Sample outputs on Wikipedia generation.
  }\label{tab:wikigen-sample-2}
\end{figure*}

\begin{figure*}[th]
    \centering
	\begin{tabular}{|p{140mm}|}
    \hline
	\textbf{Topic}: written-domain language modeling for automatic speech recognition .  \\
	\hline
	\textbf{Human:} language modeling for automatic speech recognition systems has been traditionally in the verbal domain . in this paper , we present finite-state modeling techniques that we developed for language modeling in the written domain . the first finite-state modeling techniques we describe is for the verbalization of written-domain vocabulary items , which include lexical and non-lexical entities . the second finite-state modeling techniques is the decomposition -- recomposition approach to address the out-of-vocabulary and the data sparsity problems with non-lexical entities such as urls , e-mail addresses , phone numbers , and dollar amounts . we evaluate the proposed written-domain language modeling approaches on a very large vocabulary speech recognition system for en-glish . we show that the written-domain language modeling approaches improves the speech recognition and the asr transcript rendering accuracy in the written domain over a baseline system using a verbal-domain language model . in addition , the written-domain language modeling approaches is much simpler since written-domain language modeling approaches does not require complex and error-prone text normalization and denormalization rules , which are generally required for verbal-domain language modeling . \\
	\rowcolor{lightgray!30}
	\textbf{Our model:} in this paper , we propose an approach to automatic speech recognition systems based on verbal verbal domain verbal finite-state finite-state modeling . our approach is motivated by the observation that a modeling language can be represented in a written written written in a verbalization of written-domain vocabulary items vocabulary of verbalization of written-domain vocabulary items vocabulary . this is achieved by introducing a recomposition recomposition -- approach to represent the non-lexical and lexical entities . we also present a approach to learn a set of decomposition recomposition approach . this approach can be applied to sparsity sparsity problems in the written domain . the approach is tested on the task of e-mail e-mail addresses . it is shown that it is possible to achieve accuracy comparable to that of a state-of-the-art approach . in addition , the \\
	\textbf{Our model (Oracle Plan):} in this paper , we present automatic speech recognition systems based on verbal verbal domain . our approach is based on finite-state modeling of the modeling language in the form of a domain written domain . the finite-state finite-state finite-state techniques is designed to capture the verbalization of written-domain vocabulary items verbalization of written-domain vocabulary items verbalization of written-domain vocabulary items of verbalization of written-domain vocabulary items items , and exploit the relationship between entities and lexical entities . a approach based on recomposition recomposition recomposition approach , called out-of-vocabulary , is applied to the problems of out-of-vocabulary out-of-vocabulary out-of-vocabulary problems , and a approach to e-mail e-mail e-mail addresses sparsity problems with entities , entities , urls , urls and urls , and urls . experimental results show that the proposed recomposition -- -- recomposition -- recomposition recomposition -- approach can be effective for addresses e-mail addresses and asr transcript rendering rendering rendering accuracy . \\
	\hline
	\end{tabular}
	\caption{
	Sample outputs on abstract generation.
  }\label{tab:absgen-sample-1}
\end{figure*}

\begin{figure*}[th]
    \centering
	\begin{tabular}{|p{140mm}|}
    \hline
	\textbf{Topic}:  optimal linear feature transformations for semi-continuous hidden markov models . \\
	\hline
	\textbf{Human:} linear discriminant or karhunen-lo eve transforms are established techniques for mapping features into a lower dimensional subspace . this paper introduces a uniform statistical framework , where the computation of the optimal feature reduction is formalized as a maximum-likelihood estimation problem . the experimental evaluation of this suggested extension of linear selection methods shows a slight improvement of the recognition accuracy. \\
	\rowcolor{lightgray!30}
	\textbf{Our model:} in this paper , we address the problem of estimating the number of mapping features in a lower dimensional lower dimensional space . we propose a new statistical framework based on reduction of optimal feature optimal optimal optimal feature reduction . we show that the proposed method can be applied to a variety of mapping mapping and features , and demonstrate its superior performance in terms of both speed and recognition recognition recognition performance , as well as improving the accuracy of the proposed method to the best of our knowledge this is the first attempt to improve the accuracy of the mapping mapping by up to a factor of three orders of \\
	\textbf{Our model (Oracle Plan):} in this paper , we address the problem of estimating the number of mapping features in a lower dimensional lower dimensional space , given a set of dimensional points of a given matrix , and a sparse set of points in a large number of variables.\\
	\hline
	\end{tabular}
	\caption{
	Sample outputs on abstract generation.
  }\label{tab:absgen-sample-2}
\end{figure*}